\documentclass[conference,a4paper]{IEEEtran}
\IEEEoverridecommandlockouts

\usepackage{cite}
\usepackage{amsmath,amssymb,amsfonts}
\usepackage{algorithmic}
\usepackage{graphicx}
\usepackage{textcomp}
\usepackage{xcolor}
\usepackage{hyperref}
\usepackage{tikz}
\usetikzlibrary{calc,shapes,arrows}
\usepackage[mode=buildnew]{standalone}%

\usepackage[OT1]{fontenc}

\usepackage{tabularx,booktabs}
\newcolumntype{C}{>{\centering\arraybackslash}X}
\def\BibTeX{\textrm{B\kern-.05em\textsc{i\kern-.025em b}\kern-.08em
T\kern-.1667em\lower.7ex\hbox{E}\kern-.125emX}}

\usepackage[acronym]{glossaries}
\usepackage{glossaries-extra}
\glsdisablehyper
\usepackage{csquotes}
\usepackage{cleveref}
\setabbreviationstyle[acronym]{long-short}

\global\marginparsep=0pt
\global\marginparwidth=47pt
\global\marginparpush=10pt
\newcommand{\sthu}[1]{\marginpar{\tiny\textcolor{red}{StHu: #1}}}
\newcommand{\sis}[1]{\marginpar{\tiny\textcolor{blue}{SiS: #1}}}
\newcommand{\uram}[1]{\marginpar{\tiny\textcolor{green}{UraM: #1}}}

\renewcommand{\uram}[1]{}
\renewcommand{\sthu}[1]{}
\renewcommand{\sis}[1]{}

\newacronym{opcua}{OPC~UA}{Open Platform Communications Unified Architecture}
\newacronym{rl}{RL}{Reinforcement Learning}
\newacronym{ai}{AI}{Artificial Intelligence}
\newacronym{ml}{ML}{Machine Learning}
\newacronym{dt}{DT}{Digital Twin}
\newacronym{im}{IM}{Information Model}
\newacronym{gnns}{GNNs}{Graph Neural Networks}
\newacronym{lstms}{LSTMs}{Long short-term Memories}
\newacronym{ppo}{PPO}{Proximal Policy Optimization}
\newacronym{iot}{IoT}{Internet of Things}
\newacronym{i4}{I4.0}{Industry 4.0}
\newacronym{ot}{OT}{Operational Technology}
\newacronym{mtp}{MTP}{Module Type Package}
\newacronym{ql}{QL}{Q-Learning}
\newacronym{dql}{D-QL}{Deep Q-Learning}
\newacronym{drl}{DRL}{Deep Reinforcement Learning}
\newacronym{dnn}{DNN}{Deep Neural Network}
\newacronym{mdp}{MDP}{Markov Decision Process}

\newcommand{\institute}{Salzburg University of Applied Sciences}
\newcommand{\department}{Josef Ressel Centre for Intelligent and Secure Industrial Automation}

\begin{document}
\newboolean{anonymized}

\setboolean{anonymized}{false}

\ifthenelse{\boolean{anonymized}}
{
    \title{A Mini Review on the utilization of Reinforcement Learning with
        OPC~UA ~\ifdefined\commitsha \textbf{Commit: \commitsha}\fi
    }

    \author{
        \IEEEauthorblockN{Author 1, Author 2, Author 3}
        \IEEEauthorblockA{Research Unit \\ University\\
            ZIP City, Country\\
        \{ firstname.lastname \}@mail.com}
    }
}
{
    \title{A Mini Review on the utilization of Reinforcement Learning with
        OPC~UA\thanks{
        \textcopyright 2023 IEEE. Personal use of this material is permitted.
        Permission from IEEE must be obtained for all other uses, in any
        current or future media, including reprinting/republishing this
        material for advertising or promotional purposes, creating new
        collective works, for resale or redistribution to servers or lists,
        or reuse of any copyrighted component of this work in other works.
        The final authenticated version is available online at:
        \url{https://doi.org/10.1109/INDIN51400.2023.10218289}.}
    }

    \author{
        \IEEEauthorblockN{Simon Schindler, Martin Uray, Stefan Huber}
        \IEEEauthorblockA{\department\\\institute\\
            5412 Puch/Hallein, Austria\\
        \{ simon.schindler, martin.uray, stefan.huber \}@fh-salzburg.ac.at}
    }
}

\maketitle	      %
\begin{abstract}
    \gls{rl} is a powerful machine learning paradigm that has been applied
    in various fields such
    as robotics, natural language processing and game playing
    achieving state-of-the-art results.
    Targeted to solve sequential decision making problems, it is by design able
    to learn from experience and therefore adapt to
    changing dynamic environments.
    These capabilities make it a prime candidate for controlling and optimizing
    complex processes in industry.
    The key to fully exploiting this potential is the seamless integration of
    \gls{rl} into existing industrial systems.
    The industrial communication standard \gls{opcua} could bridge this gap.

    However, since \gls{rl} and \gls{opcua} are from different fields, there
    is a need for researchers to bridge the gap between the two technologies.
    This work serves to bridge this gap by providing a brief technical
    overview of both technologies and carrying out a semi-exhaustive
    literature review to gain insights on how \gls{rl} and \gls{opcua} are
    applied in combination.

    With this survey, three main research topics have been identified,
    following the intersection of \gls{rl} with \gls{opcua}.
    The results of the literature review show that \gls{rl} is a promising
    technology for the control and optimization of industrial processes, but
    does not yet have the necessary standardized interfaces to be deployed in
    real-world scenarios with reasonably low effort.
\end{abstract}

\begin{IEEEkeywords}
    OPC~UA, Reinforcement Learning, Survey
\end{IEEEkeywords}

\section{Introduction}
\label{sec:introduction}

The fourth industrial
revolution aims to improve
efficiency and productivity, while reducing costs and downtime of
production systems, by establishing interconnectivity and intercommunication
of all relevant participants in the production processes.
Successful and long-term sustainable integration of participants like \gls{iot}
devices, \gls{ml} algorithms and \glspl{dt} into production processes presents
a significant challenge for the manufacturing industry.
To enable systematic research and implementation strategies,
the four design principles for the \gls{i4} of Hermann et
al.~\cite{Hermann2016}
should be adhered to:
\begin{enumerate}
      \item Interconnection: Secure communication of all participants
            relevant to
            the production process, such as machines, devices, sensors and
            people.
      \item Information transparency: Context-aware information presentation
            assures interpretability.
      \item Decentralized decisions: Enabled by the interconnection of all
            devices and a given semantic for the provided information
            participants can
            make decentralized decisions.
      \item Technical assistance: Technical assistance aggregating and
            visualizing relevant information is substantial for
            the ongoing shift of the role of humans from machine operators to
            strategic decision-makers in manufacturing.
\end{enumerate}

\textit{\gls{opcua}} is a common industrial communication standard
that provides reliable and secure end-to-end
communication of data and events between cross-platform devices
and software applications~\cite{opc_website}.
It is used in a variety of industries and acts as a de facto standard
communication and information modeling protocol within \gls{ot}.
It is an enabling technology for flexible,
adaptive and transparent production
systems and has a wide range of use cases of which an overview is given
in~\cite{Schleipen2016}.
With its standardized data model~\cite{opc_reference}, it does not
only provide a framework for the communication of data and events,
but also enables the modeling of the
entire industrial information network in an object-oriented way.
Therefore, \gls{opcua} satisfies the first two design principles of
\gls{i4}
by providing a secure intercommunication platform (1.~interconnection) and
by giving communicated
information a semantic that can be interpreted by all participants
(2.~information transparency).

\textit{\gls{rl}} is a popular \gls{ml} technique that has been
widely
applied in various fields, including robotics~\cite{Kober2013},
natural language processing~\cite{Cetina2022,Luketina2019},
and game playing at even superhuman levels~\cite{atari,alphago}.
It is able to learn on the fly, adapt to ever-changing
environments and is specifically well suited for solving sequential
decision-making
problems, such as the control of a production system.
Its possible applications in an industrial environment are manifold
as outlined by the extensive review on \gls{rl} in industrial process control
provided by Nian in~\cite{Nian2020}.
In combination with \gls{opcua}, it could potentially suffice the third and
fourth
design needs of Herrman et al.\ by acting as a technical
assistant for human decision makers or even as a decision maker itself.
\gls{opcua} is specifically well suited to be used with \gls{rl} as it
fulfills two essential needs of a \gls{rl} environment: It provides access to
information relevant to the decision-making process and it enables the agent
to act on the environment.

The combination of both technologies, \gls{opcua} and \gls{rl}, could in theory
enable continuously self-optimizing control and operation of industrial systems
and systems-of-systems of large scale resulting in a significant increase in
efficiency and productivity.

To the best of our knowledge, there exists no overview of scientific literature
that utilizes \gls{opcua} in combination with \gls{rl}.
Hence, this work would be the first contribution in this area.
Through that, we aim to open up
this field of research to both researchers and practitioners of the
manufacturing and
\gls{ml} domain.

To achieve that we first provide a short
introduction to the technical terminology of both \gls{opcua} and \gls{rl}
and then present a selection of research articles that utilize \gls{opcua} in
combination with \gls{rl}.
While this field is not yet to be recognized as a much successful application
domain, we show that there are various research articles around, that
demonstrate the rich interplay of both, \gls{rl} and \gls{opcua}.

\section{Terminology and technical background}
\label{sec:sec_terminology}
\gls{opcua} and \gls{rl} are nested within
different domains of computer science.
In order to make the discussion of both topics more accessible to the
reader of
this article, who may belong to
just one of these fields, this section briefly provides some
basic information and terminology on both backgrounds.

\subsection{Reinforcement Learning}\label{subsec:terminology_rl}
\gls{rl} is a \gls{ml} paradigm, that explicitly does not need any
supervisory data. Its approach of learning towards optimal decisions
differs from supervised
learning in that sense that \gls{rl} agents learn from experiences through
interactions with an environment rather than from labeled data.

\gls{rl} is specifically designed to solve
sequential decision
making problems and can deal with high-dimensional state and action spaces.
Therefore it is very well suited for the control of industrial
processes, which often provide a vast amount of data to consider (the state
space) and have a
high number of degrees of freedom which actions to take (the action space).

\begin{figure}
    \centering
        \begin{tikzpicture}
        [auto,
        node distance=1.5cm,
        block/.style = {draw, fill=white, rectangle, minimum height=1cm,
        minimum width=2.3cm},
        ]

        \node [block] (agent) {Agent};
        \node [block, below of=agent] (environment) {\small Environment};

        \draw [->] (agent.east)	-- ($ (agent.east) + (1.25,0) $)  |- (
        environment.east);
        \draw [->] (environment.170) -- ($ (environment.170) - (1.0,0) $)
        |- (agent.190);

        \draw [->] (environment.190) -- ($ (environment.190) - (1.5,0) $)
        |- (agent.170);

        \draw [dotted] (-1.8,-1) -- (-1.8,-2);

        \node[align=left,rotate=90] at ($ (environment.east) + (1.1,0.85) $)
            {\scriptsize Action $a_i$ } ;

        \node[align=left,rotate=90] at ($ (environment) + (-2.8,0.85) $)
            {\scriptsize State $s_i$ } ;
        \node[align=left,rotate=90] at ($ (environment) + (-2.3,0.85) $)
            {\scriptsize Reward $r_i$ } ;
        \node at ($ (environment.190) + (-.3,0.15) $) {\scriptsize $s_{i+1}$ };
        \node at ($ (environment.170) + (-.3,0.15) $) {\scriptsize $r_{i+1}$ };

    \end{tikzpicture}
    \caption{The \gls{rl} agent-environment interaction according Sutton and
        Barto~\cite{Sutton2018}.}
    \label{fig:rl}
\end{figure}
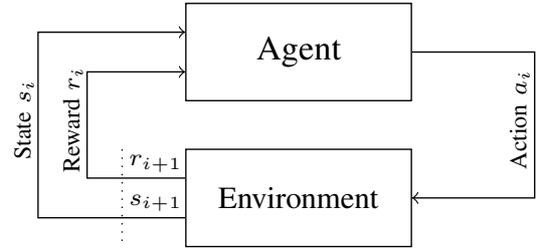

The \gls{rl} concept, as illustrated in Figure~\ref{fig:rl}, is systematically
very similar
to control theory. The analog to the controller would be the agent, the
environment would be the plant and the reward would be the control error.
But in contrast to classical control theory, the agent is not given a model (with
some exceptions) of the plant,
but has to learn it from experience. Since its model of the environment changes
with learning iterations, its performance, dependent on the quality of the
model,
is not static as in classic control theory,
but dynamic.

This type of learning is very similar to the learning process of humans,
where one is rewarded or punished for decisions taken.
Further decisions are impacted by the consequences in
the past.

Many \gls{rl} problems can be formulated as \glspl{mdp}.
Let $S$ be the set of all possible states of the environment and $A$ the set of
all possible actions.
In a \gls{mdp}
the transitions from one state $s_i$ to the next $s_{i+1}$ is
stochastic, controlled by the taken action $a_i$.
During training of an agent, given a state $s_i \in S$, at each time step $i$,
the agent picks an action $a_i \in A$ that leads to a reward $r_{i+1} \in
    \mathbb{R}$ and a successor state $s_{i+1} \in S$.
The goal of the agent is to max.\ the (discounted) cumulative reward
$G_t = \sum_{i=t+1}^{T} \gamma^{i-t-1} \cdot r_i$, where $t$ is the current
time step, $T$ the optimization horizon and
$\gamma \in (0,1)$ is a discount factor.
This is achieved by the adaption of the agents' policy $\pi: S \rightarrow A$,
a mapping from states to actions.

For a more detailed introduction to \gls{rl}, the reader may be referred to
Sutton and Barto~\cite{Sutton2018}.

\subsection{OPC UA}\label{subsec:terminology_opcua}
The \gls{opcua} is a standard by the IEC~\cite{opc_reference}
released in 2008
that specifies a cross-platform data exchange format for sensory- and
machine data.
In comparison to other standards, with \gls{opcua} not only the data is
exchanged, also the semantics of the data and relationships between network
participants are included. This makes
\gls{opcua} as a communication architecture specifically well suited for the
use of \gls{ml} methods such as \gls{rl} in a manufacturing environment. This extra information can be at least beneficial but even substantial for
successfully
applying \gls{ml}.

For the purpose of semantically describing the data and the participants
interrelations, so-called \glsfirstplural{im} are utilized.
Within such a \gls{im}, the definition of the structure, data types, and
relationships between the various information resources that are available
for communication are modeled.
This framework utilizes flexibility and extensibility, by enabling
dynamic adaptions, e.g.\ adding new \textit{nodes}.
The nodes represent, among other things, objects, variables or
methods that are accessible through the network.
Each node is uniquely identifiable by its
\textit{NodeID} and described by its \textit{NodeClass} defining its
properties and capabilities.
Furthermore, nodes can have \textit{references} to one another,
representing their relationships and the ways in which they can be accessed.
For example, \textit{object nodes} are
physical or abstract objects within the system and can have properties in
the
form of
\textit{variable nodes} and callable software functions accessible through
\textit{method nodes}.

Communication in \gls{opcua} is always established between a
\textit{client}
and a
\textit{server}, both of which a system can possess multiple instances.
The flexibility of the
architecture allows for each client to be concurrently connected to
multiple
servers and vice versa.
The server provides the client with an interface to information by granting
access to nodes
through the use of its so-called address space.
This information, once obtained, can be
interpreted by the client through the utilization of the corresponding
\gls{im}.

In short \gls{opcua} models the information and the relationships
of participants in an industrial environment and provides the possibility
to interact with the participants. This makes it a suited candidate to enable \gls{rl} in an industrial environment, since
\gls{rl} needs both information about the environment and
a way of interacting with it.

\section{Methodology}
\label{sec:methodology}
The research methodology used for this semi-exhaustive literature review
is a \textit{reproducible} four-step procedure.
This method is
motivated by Tschuchnig et al.~\cite{Tschuchnig2022} and based on
Randolph~\cite{Randolph2019}.
The problem defined for this review is the evaluation of the application of
\gls{rl} with \gls{opcua}.

For proper search queries, the terms \enquote{reinforcement~learning}
and \enquote{opc~ua} were identified.
No further keywords were considered to be included.

For data collection, the search engine \textit{Google Scholar} was chosen.
This search engine enables to specifically formulate a search query using
boolean operators.
Additionally, quotation marks can be used to indicate, that the containing
keywords must be included within the results.
Using these properties, the search query
    \enquote{reinforcement~learning} AND
    \enquote{opc~ua}
was used.
Giving $10$ results per page, we limited the data collection to the results
of the $6$ first pages, i.e.\ resulting in $60$ references of different
kind.

For further processing, a set of criteria was defined:
In order to deliver a qualitative review, for this work only conference- and
journal publications were taken into account.\footnote{We are aware that not
    all conference- and journal publications may be peer-reviewed.}
Preprints, presentations, reports, or commercial information were excluded.
Furthermore, this review takes all publications into account, that were
published before 15th of February 2023, with no restrictions on a lower bound.
The language of the publication had to be English and a full-text version had
to be available for further steps.

The last filtering step was to read the full-text, to find out if the work is
in the context of our review's research interest.
Publication just naming the methodologies, e.g.\ in a related work section or
discussion, were excluded.

As a final step, the remaining literature was analyzed and were grouped into
manually identified clusters.

\section{Results}
\label{sec:results}
The result of the semi-exhaustive literature review exposed a number of $17$
conference- and journal papers.
In the following, the three identified categories are listed and elaborated.

\subsection{Cluster 1: Industrial Applications of RL utilizing
    OPC UA for communication} \label{subsec:res_cluster1}
The first cluster consists of ten papers.
All of them applied \gls{rl}
in an industrial setting and relied on \gls{opcua} just for communication.
The type of applications is diverse.
These range from an agent learning to play table football using
industrial drives~\cite{Blasi2021, Rohrer2021} to optimizing energy costs
of a production plant~\cite{Bakakeu2018,Schmidl2020,Schmidl2021}.
They all did not directly address the combination of \gls{rl} and \gls{opcua},
but rather focused on the application of \gls{rl} in an industrial setting,
while applying \gls{opcua} just for communication.
Hence, they are not described in further detail in this work.

An interesting observation in this cluster is the predominant
utilization of virtual simulations for pre-training.
The papers gathered in cluster 1 are listed in table~\ref{tab:cat1}.
There, the results are summarized highlighting their learning methods, the
use of virtual simulations for pre-deployment training, and the specific
applications applied on.

\begin{table*}
    \begin{tabularx}{\textwidth}{@{} l p{4.5cm} @{} c @{} *{1}{X}}
        \toprule
        Ref.\                        & \gls{rl}-method
                                         & use simulation~~
                                         &
        summary
        \\
        \midrule
        \,\cite{Schmidl2020,Schmidl2021} & \gls{ql}
                                         & $\checkmark$
                                         & 
        energy reduction of production plants by dynamic hibernation of
        sub-systems depending on current and future estimated production load
        \\
        \,\cite{Burggraf2022}            & \gls{ql}
                                         & $\times$
                                         &
        process time optimization in an assembly process

        \\
        \,\cite{Xia2021}                 & \gls{dql}
                                         & $\checkmark$
                                         &
        evaluation of \glspl{dt} for industrial \gls{rl}

        \\
        \,\cite{Dogru2022}               & On- and Offline Policy Gradient
        Methods with
        DNNs                             & $\times$
                                         & automated PID controller tuning in
        an industrial
        environment

        \\
        \,\cite{Bakakeu2018}             & \gls{dql}
                                         & $\times$
                                         &
        energy cost
        optimization through adaptive utilization of production systems
        depending on energy prices

        \\
        \,\cite{Blasi2021}               & Actor-Critic Policy Gradient Method
        with DNNs                        &
        $\checkmark$                       & simulation based deep-\gls{rl}
        to
        play the attacker at table
        football using industrial drives

        \\
        \,\cite{Rohrer2021}              & \gls{dql}
                                         & $\checkmark$
                                         &
        simulation based
        deep-\gls{rl} to play the goalkeeper at table football using
        industrial drives

        \\
        \,\cite{Dobrescu2020}            & Deep-\gls{rl} (no further details
        provided)                        &
        $\checkmark$                       & hard- and software in the loop
        assessment of an industrial
        control architecture

        \\
        \,\cite{Abdoune2022}             & \gls{ppo}
                                         & $\checkmark$
                                         &
        case
        study of the integration of \gls{rl} into the life cycle of
        industrial \glspl{dt}

        \\
        \bottomrule
    \end{tabularx}
    \caption{Cluster 1: industrial applications of \gls{rl} just relying on
        \gls{opcua} for communication}
    \label{tab:cat1}
\end{table*}

\subsection{Cluster 2: Architectures for the integration of RL into
    industrial environments utilizing OPC UA} \label{subsec:res_cluster2}
To enable the switch from the simulation environment to the actual industrial
environment, the interface between the agent and the affected systems should
not differ too much from the simulation.
Using \gls{opcua} for communication between the agent and the environment
already in simulation facilitates the switch, but still requires some
implementation effort, caused by a lack of a standardized interface for
\gls{rl} within \gls{opcua}.
This lack of architectural standardization is addressed by the second cluster
of papers analyzed, listed in table~\ref{tab:cat2}.

Grothoff et al.~\cite{Grothoff2021} propose a mapping of states of standardized
state
machines to generic interactions of \gls{rl} algorithms for learning
and inference phases.
They showcase a proof of concept in the form of a simulation of a coil
transport system for cold rolling steel mills.

An architecture for a modular \gls{rl} environment generator is proposed by
Csiszar et al.~\cite{Csiszar2021}.
This generator operates as a  configurable adapter between the \gls{rl} agent
and industrial communication protocols such as \gls{opcua}.
Their proposed method especially targets practitioners of the domain of
production engineering,
with no prerequisites in software engineering, to enable easy integration of
\gls{rl}
algorithms into industrial environments for non domain experts.
However, this work is only of conceptual nature and
no implementation of such a generator has been published.

In~\cite{Schaefer2022}, the authors propose an \gls{opcua} \gls{im} to deploy
\gls{rl} algorithms to industrial environments.
Therefore, \gls{opcua} nodes are extended by sensor and action properties. This
enables automatic generation of the observation- and action space and the
deployment of \gls{rl} agents without an engineering effort.
Using this architecture, a \gls{rl} agent has been successfully deployed to a
model example using a Hardware-in-the-Loop simulation.

Gracia et al.\ \cite{Gracia2022} created a framework for the integration of
robotics, in terms of hardware and control software into industrial
environments.
The communication is based on \gls{opcua}.
A configurable \gls{opcua} server hosts the control logic, such as \gls{rl}
algorithms for robotic components.
In turn, these are accessible as \gls{opcua} clients as well.
The attachment of a robotic component to the framework is enabled by
template-based plugins.

In~\cite{Khaydarov2022}, the authors developed a framework based on
\gls{opcua} for communication which can be used to integrate \gls{rl}
algorithms into industrial environments.
In contrast to~\cite{Gracia2022}, this is rather focused on \gls{ml} methods
than on robotics.
Their approach enables a native integration via the Python programming language
by means of common \gls{ml} libraries such as TensorFlow, PyTorch
and others.
Furthermore, their work requires all used components to be compatible
with the \gls{mtp}, a standardized description of the automation interface
for self-contained production units.

\begin{table*}
    \begin{tabularx}{\textwidth}{@{} l c @{} *{1}{X}}
        \toprule
        Ref.\              & open-source~~ & summary
        \\
        \midrule
        \,\cite{Csiszar2021}   & $\times$      & environment
        generator as
        adapter
        between interface of agent and industrial communication protocols
        \\
        \,\cite{Grothoff2021}  & $\times$      & mapping of
        standardized
        state
        machine
        states to generic interactions of \gls{rl} algorithms
        \\
        \,\cite{Schaefer2022}  & $\times$      & \gls{opcua}
        \gls{im} for
        deploying and
        exchanging the RL agent
        \\
        \,\cite{Gracia2022}    & $\checkmark$  & framework
        allowing
        integration
        of
        external hardware and software (s.a.\ a \gls{rl} agent) based on
        \gls{opcua}
        \\
        \,\cite{Khaydarov2022} & $\checkmark$  & framework
        allowing
        integration
        of
        external hardware and software (s.a.\ a \gls{rl} agent) based on
        \gls{opcua} and \gls{mtp}
        \\
        \bottomrule
    \end{tabularx}
    \caption{Cluster 2: architectures for the integration of \gls{rl}
        into industrial environments utilizing \gls{opcua}}
    \label{tab:cat2}
\end{table*}

\subsection{Cluster 3: RL applied for information inferences
    from OPC UA IMs} \label{subsec:res_cluster3}

The third cluster contains work on information inferences about \gls{opcua}
\glspl{im} by embedding them in knowledge graphs and applying \gls{rl} with
\gls{gnns}.
Knowledge graphs are semantic networks
that can be represented in \gls{opcua} through information modeling.
The cluster is listed in table~\ref{tab:cat3}.

Bakakeu et al.\cite{Bakakeu2020} train a \gls{rl} agent to be capable of
inferring information from semantically incomplete \glspl{im}.
Here, the aim is to discover missing relationships between nodes and the
application of consistency checks.
This was achieved by constructing multi-hop relationship paths along the
embedding vector space of the knowledge graph.
The knowledge graph represents the \gls{im} by training a \gls{rl} agent,
based on \gls{lstms} with \gls{ppo}, with the aim to predict the
relationships between the given entities.

A similar approach is taken by Zheng et al.\ in~\cite{Zheng2021}.
They establish a manufacturing system capable of a semantics-based
solution search on the established knowledge graph, that
enables the organization of available
on-site manufacturing resources.
To this end, they train multiple agents on an embedding of a knowledge graph.
In this case, this graph is not constructed from the \gls{opcua} \glspl{im}
only but on information on the relationships from multiple sources.
Since their method builds on the initially constructed knowledge graph, they
highlight the challenge of dealing with ever-changing dynamic environments as
an open research question in this area.

\begin{table*}
    \begin{tabularx}{\textwidth}{@{} l *{2}{X}}
        \toprule
        Ref.\                           & \gls{rl}-method
                                            & summary
        \\
        \midrule
        \,\cite{Bakakeu2020}                & \gls{ppo}
                                            & \gls{rl} for
        reasoning on semantically incomplete \gls{opcua} \glspl{im} for the
        discovery of missing relations between the entities
        \\
        \,\cite{Zheng2021}                  & multi-agent independent learning
        with a soft
        actor critic policy gradient method & knowledge graph based multi-agent
        \gls{rl} for self-(configuration, optimization, adjustment) of a
        manufacturing network
        \\
        \bottomrule
    \end{tabularx}
    \caption{Cluster 3: \gls{rl} applied for information inferences from
        \gls{opcua} \glspl{im}}
    \label{tab:cat3}
\end{table*}

\section{Discussion}
\label{sec:discussion}
The first cluster exposes the variety of opportunities for the application of
\gls{rl} in industrial environments, while building on \gls{opcua} as a
communication platform.
In eight out of these ten papers, virtual simulations of the environment for
parts of the training process were used.
The utilization of simulations enables faster training due to a separation from
the production environment during the initial training phase and the
possibility of parallelization.
This separation from real machines also prevents the system from causing
costly downtimes, as well as the introduction of possible hazardous
situations during random exploration.
Furthermore, this approach enables a much more convenient form of
experimentation, since	different algorithms and their hyperparameters can be
tested without interaction with the real-world.
The resulting number of publications in this cluster shows, that this topic is
of general interest and of importance.
In many cases, the simulations of the real-world environment must be enabled
and communication interfaces in the simulation must be consistent with
reality.
This seems to be a key factor for successful deployment, which strongly
points in the direction of \glspl{dt} and \gls{opcua}.

The great potential of utilizing \gls{rl} in industrial environments is
limited by the lack of a standardized integration within \gls{opcua}.
In order to facilitate the application of \gls{rl} in industrial environments,
this bottleneck needs to be addressed.
Still, only five publications have been found that address this issue to at
least some extent.
Here, much more research and actual practical considerations are needed in the
future.
The concept of Csiszar et al.\ for a \gls{rl}-environment generator for
integrating \gls{rl} into industrial networks~\cite{Csiszar2021} would be a
major contribution to this field, if it did not lack an implementation.
The work of Schäfer et al., which proposed a \gls{opcua} \gls{im} for
\gls{rl}~\cite{Schaefer2022} could be further developed to provide a solid
basis for future official standardization.
The two implemented
frameworks for integrating \gls{rl} with
\gls{opcua}~\cite{Gracia2022,Khaydarov2022} are promising to significantly
reduce the programming effort for successful integration of an
already pretrained agent,
even though they are
not fully tailored to \gls{rl}.

The work in Cluster 3 shows that \gls{rl} is a suitable methodology among other
approaches for working with knowledge graphs created from \gls{opcua}
\glspl{im}.
But still, they are not yet able to adapt to changing environments.
Extending these systems to provide adaptivity, would be a great improvement,
since this could pave the way for
self-optimizing and self-configuring industrial networks.

This work is the first survey to highlight aspects of \gls{ml} in the
context \gls{opcua}.
For further work, this survey could be extended to include other \gls{ml}
methods for a more holistic overview.
\section{Conclusion}
\label{sec:conclusion}
\gls{opcua} models the information and relationships of participants in an
industrial environment and provides a way to interact with the participants.
This makes it a perfect candidate to enable \gls{rl} in an industrial
environment, since \gls{rl} requires both information about the environment and
a way to interact with it.

This work provides an overview of the current state of research
on the application of \gls{rl} with \gls{opcua} by means of a
systematic semi-exhaustive literature review.

There have been found three main clusters in which the papers can be
grouped.
Papers of the first cluster focus on using \gls{rl} to control and
optimize industrial processes,
including production lines and energy systems.
A trend that can be observed in this cluster is the use of simulation
environments and \glspl{dt}, since training
\gls{rl} agents in cost and safety-sensitive environments
can be unsuitable.
\gls{opcua} can be even used in the simulations to closely resemble
the real-world industrial environment, in terms of behavior on interaction
with the agent and interfacing
by means of communication protocols and standards.

The integration of \gls{rl}
algorithms with \gls{opcua} is a significant challenge addressed by the papers
in the second cluster found,
since there does not exist a standardized framework for installing a \gls{rl}
environment in an industrial setting.
Worked on by only a few authors there is much room for research left in
this area to provide a solid base for future standardization processes.
The lack of open-source implementations of the proposed
frameworks is a major drawback of the papers in this cluster.
Research in this subcategory also publishing implementations would be most
beneficial to the whole field,
since standardized and therefore faster integration of \gls{rl} into
industrial environments would enable and encourage research in the other
subcategories.

The last group of papers considered the use of \gls{rl}
for information inference from \gls{opcua} \glspl{im} by embedding them in
knowledge graphs.
This has been explored by some authors~\cite{Bakakeu2020,Zheng2021}.
They conclude that these techniques could be useful to enable
self-organizing
manufacturing systems in the future,
while still addressing fundamental problems when dealing with dynamically
changing industrial environments opening the field up for further research.

In conclusion, the combination of \gls{rl} and \gls{opcua} is a promising
area of research and holds significant potential for improving
industrial processes.
However, the beforehand outlined challenges need to be addressed.

\bibliographystyle{IEEEtranS}
\bibliography{export}

\end{document}